\documentclass[conference]{IEEEtran}
\IEEEoverridecommandlockouts

\usepackage{cite}
\usepackage{amsmath,amssymb,amsfonts}
\usepackage{graphicx}
\usepackage{textcomp}
\usepackage{xcolor}
\usepackage[T1]{fontenc}
\usepackage{flushend}
\usepackage{multirow}
\usepackage{multicol}
\usepackage{array}
\usepackage{url}
\usepackage{booktabs}
\usepackage{algorithm}%
\usepackage{algorithmicx}%
\usepackage{algpseudocode}%
\usepackage{subcaption}
\usepackage[dvipsnames]{xcolor}

\usepackage{fancyhdr}
\def\BibTeX{{\rm B\kern-.05em{\sc i\kern-.025em b}\kern-.08em
    T\kern-.1667em\lower.7ex\hbox{E}\kern-.125emX}}
\begin{document}

\newcommand\glebnotes[1]{\textcolor{blue}{(Gleb Note: #1)}}
\newcommand\ahmednotes[1]{\textcolor{Aquamarine}{#1}}
\definecolor{mygreen}{RGB}{61, 120, 56}
\newcommand\marcnotes[1]{\textcolor{mygreen}{#1}}
\definecolor{lightgrey}{RGB}{100, 100, 100}
\newcommand\lighter[1]{\textcolor{lightgrey}{#1}}

\title{Federated Distillation on Edge Devices:\\ Efficient Client-Side Filtering for Non-IID Data\\
}

\author{\IEEEauthorblockN{Ahmed Mujtaba\IEEEauthorrefmark{1}, Gleb Radchenko\IEEEauthorrefmark{1}\IEEEauthorrefmark{3}, Radu Prodan\IEEEauthorrefmark{2}\IEEEauthorrefmark{5} and Marc Masana\IEEEauthorrefmark{4}\IEEEauthorrefmark{3}}
\IEEEauthorblockA{\IEEEauthorrefmark{1}Embedded Systems Division, Silicon Austria Labs, Graz, Austria}
\IEEEauthorblockA{\IEEEauthorrefmark{2}Department of Computer Science, University of Innsbruck, Austria}
\IEEEauthorblockA{\IEEEauthorrefmark{5}Institute of Information Technology, University of Klagenfurt, Austria}
\IEEEauthorblockA{\IEEEauthorrefmark{3}TU-Graz SAL DES Lab, Silicon Austria Labs, Graz, Austria}
\IEEEauthorblockA{\IEEEauthorrefmark{4}Institute of Visual Computing, Graz University of Technology, Austria}
\IEEEauthorblockA{Email: \{ahmed.mujtaba, gleb.radchenko\}@silicon-austria.com, radu.prodan@uibk.ac.at, mmasana@tugraz.at}
}

\maketitle
\thispagestyle{fancy}

\begin{abstract}
Federated distillation has emerged as a promising collaborative machine learning approach, offering enhanced privacy protection and reduced communication compared to traditional federated learning by exchanging model outputs (soft logits) rather than full model parameters.  However, existing methods employ complex selective knowledge-sharing strategies that require clients to identify in-distribution proxy data through computationally expensive statistical density ratio estimators. Additionally, server-side filtering of ambiguous knowledge introduces latency to the process. To address these challenges, we propose a robust, resource-efficient EdgeFD method that reduces the complexity of the client-side density ratio estimation and removes the need for server-side filtering. EdgeFD introduces an efficient KMeans-based density ratio estimator for effectively filtering both in-distribution and out-of-distribution proxy data on clients, significantly improving the quality of knowledge sharing. We evaluate EdgeFD across diverse practical scenarios, including strong non-IID, weak non-IID, and IID data distributions on clients, without requiring a pre-trained teacher model on the server for knowledge distillation. Experimental results demonstrate that EdgeFD outperforms state-of-the-art methods, consistently achieving accuracy levels close to IID scenarios even under heterogeneous and challenging conditions. The significantly reduced computational overhead of the KMeans-based estimator is suitable for deployment on resource-constrained edge devices, thereby enhancing the scalability and real-world applicability of federated distillation. The code is available online\footnote{https://opensource.silicon-austria.com/mujtabaa/edgefd} for reproducibility. 
\end{abstract}


\section{Introduction}
\emph{Federated learning (FL)} is a branch of machine learning (ML) where multiple data-owning clients collaboratively train an ML model by updating models locally on private data and aggregating them globally~\cite{mcmahan2017communication,DS-FL,shao2024selective}. Its principal advantage is the decoupling of model training from directly accessing the raw data by
only sharing its parameters or gradients to a centralized server while the raw training data remains securely stored on local devices, enhancing privacy and potentially reducing communication~\cite{mcmahan2017communication,Khan21}.
Despite its advantages, FL faces several challenges. Firstly, the high computational resources on client devices pose significant constraints for edge devices with limited processing capabilities. Secondly, the non-independent and identically distributed~(Non-IID) data distribution across clients frequently results in imbalanced model training, impacting overall model performance~\cite{li2020federated}. Finally, the high communication cost of transmitting model gradients from edge nodes to the server during training becomes demanding for large-scale neural networks~\cite{wang2019adaptive}.

\emph{Federated distillation (FD)}, a generalization of knowledge distillation~\cite{jeong2018communication,hinton2015}, offers a practical solution to the key limitations of traditional FL. Existing FD methods in literature~\cite{qin2024knowledge} can be feature-based, parameter-based, and data-based. Our work is closely related to feature-based FD where clients share model outputs (\textit{logits}) on a publicly shared dataset (\textit{proxy data}) instead of exchanging model parameters, aggregated to refine the global model~\cite{sattler22,radchenko2024}. Other FD methods (parameter-based and data-based) commonly pose privacy concerns, are communication-expensive, or lack user personalization~\cite{qin2024knowledge}. Feature-based FD significantly reduces communication costs as the model logits are typically orders of magnitude smaller than the model gradients. Additionally, sharing model logits rather than model gradients allows system heterogeneity in which each federated device can deploy a customized ML model based on its available resources, eliminating the need for agreement on a shared model, typical in traditional FL. Additionally, feature-based FD mitigates data heterogeneity (non-IID) by combining diverse client knowledge within the proxy data, resulting in more robust and generalized models.

Despite these advantages, feature-based FD brings inherent drawbacks. It relies on proxy data whose distribution must overlap with every client for better performance. Moreover, proxy data exchange could potentially leak sensitive user information and, therefore, demands stronger privacy safeguards. Finally, state-of-the-art (SoTA) feature-based FD methods often rely on computationally expensive and memory-intensive statistical density ratio estimators~\cite{sugiyama2012density,KulSIF} to filter proxy data and transmit selected proxy logits to the server~\cite{shao2024selective}, limiting their applicability on resource-constrained devices. Additionally, server-side filtering of ambiguous knowledge can further increase the distillation overhead.

\begin{figure*}[t]
    \centering
    \begin{subfigure}[b]{0.53\textwidth}
        \includegraphics[width=\textwidth]{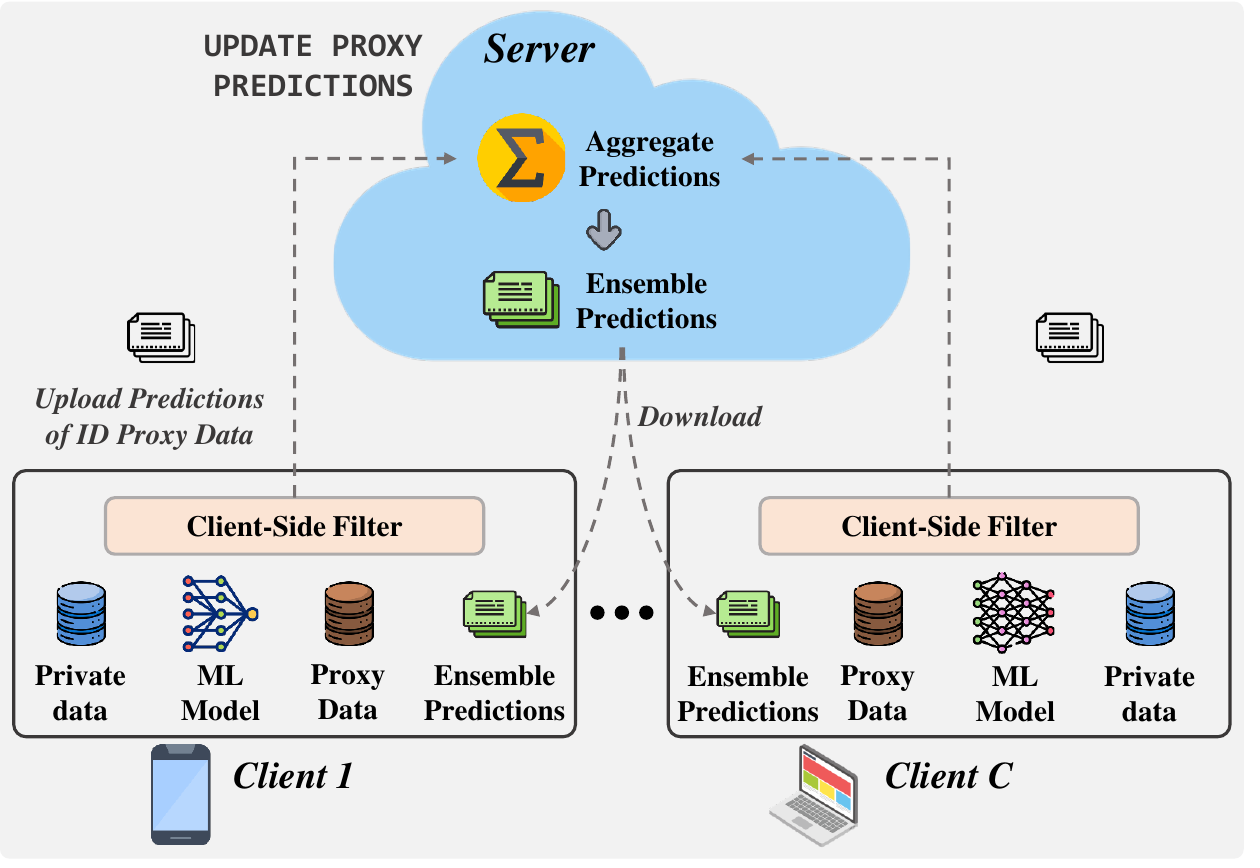}
        \caption{Global EdgeFD view.}
        \label{fig:prop_work_server}
    \end{subfigure}
    \begin{subfigure}[b]{0.44\textwidth}
        \includegraphics[width=\textwidth]{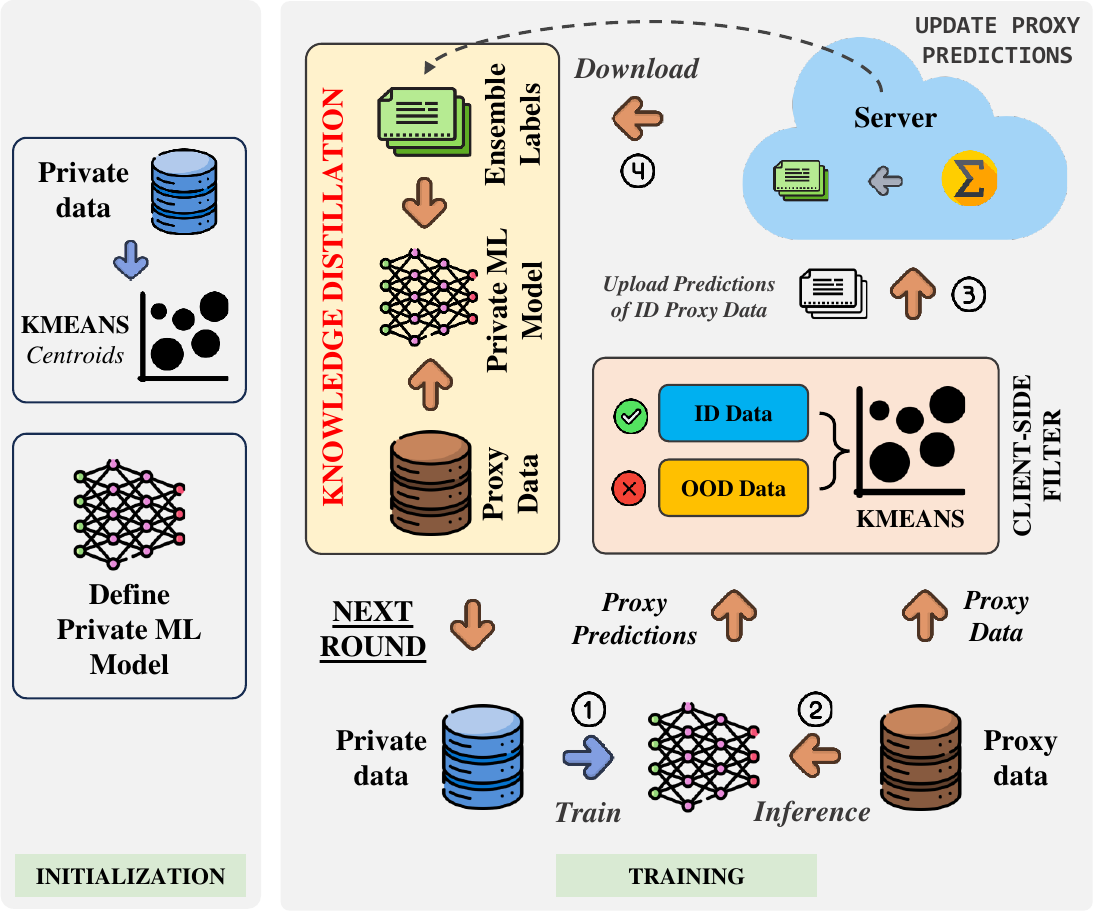}
        \caption{Client EdgeFD  view.}
        \label{fig:prop_work_client}
    \end{subfigure}
    \caption{Global and client-side workflow of EdgeFD with KMeans density ratio estimation on heterogeneous devices, customized ML models, and personalized user data \lighter{(icons courtesy of flaticon.com)}.}
    \label{fig:proposed_work}
\end{figure*}

This study introduces \emph{EdgeFD}, a lightweight edge-oriented FD method that addresses key limitations in existing FD methods \cite{shao2024selective, FedMD}. EdgeFD incorporates an innovative two-stage client-side filtering strategy powered by a resource-efficient KMeans-based density-ratio estimator (KMeans-DRE) that effectively identifies both in-distribution and out-of-distribution proxy data directly on client devices and eliminates server-side filtering. Our robust KMeans-DRE significantly reduces resource requirements compared to conventional statistical density ratio estimators by capturing data distribution through centroid positions rather than complex matrix operations. Experimental results demonstrate that EdgeFD consistently outperforms SoTA FD methods across various data distribution scenarios (strong non-IID, weak non-IID, and IID), achieving up to 98.92\% accuracy on MNIST, 88.74\% on FashionMNIST, and 86.37\% on CIFAR10 datasets. The reduced resource consumption makes EdgeFD suitable for deployment on edge devices with limited processing capabilities.

EdgeFD is immediately applicable to (i) hospital consortia that keep radiology images on-premise but can release a small de-identified subset for knowledge distillation to train a generalized model for medical applications, (ii) fleets of camera-equipped robots that upload a fraction of annotated frames through constrained wireless links and perform federated distillation to learn collaboratively \cite{radchenko2024} and (iii) enhanced driver monitoring system in automobiles to detect driver fatigue by learning collaboratively on different face features based on ethnicities and nationalities. All three settings feature heterogeneous models, strict bandwidth limits, and moderate willingness to share limited proxy data conditions.

The paper has six sections. Sec.~\ref{sec:related_works} reviews existing FD studies and their limitations. Sec.~\ref{sec:methodology} presents the EdgeFD method and the KMeans-based density-ratio estimator. Sec.~\ref{sec:experiments} covers the evaluation methodology, including data distribution, performance metrics, and datasets. Sec.~\ref{sec:discussion_and_analysis} analyzes the experimental results and the EdgeFD performance. Finally, Sec.~\ref{sec:conclusion} summarizes contributions and discusses future research.

\section{Related Works}
\label{sec:related_works}
Feature-based FD literature falls under two categories \cite{qin2024knowledge, li2024federated}: (1) proxy‑data approaches in which every participant can access the same public or collaboratively built reference proxy dataset, and (2) data‑free approaches in which clients exchange only statistics, synthetic data, or generators, eliminating the need for any shared corpus. Among data‑free approaches, baseline FD-GAN \cite{FD-GAN} avoids proxy data by exchanging only per‑label mean logits and rectifying class imbalance via a collaboratively trained GAN. FEDGEN \cite{FD-GEN} pushes this idea further, where the server learns a lightweight generator from aggregated logits broadcast and reused locally for synthetic‑feature distillation and delivering robust gains under strong heterogeneity without proxy data. FedCMD \cite{FedCMD} explores cross‑modal distillation where sensor readings act as a teacher to video‑based students inside each vehicle, and only the distilled student weights participate in FL rounds without a shared proxy. Finally, recent studies such as FKD \cite{FKD} and PLS \cite{PLS} aggregate only label‑wise statistics, avoiding proxy data entirely. These methods, though innovative, struggle to maintain performance under strong non-IID scenarios and heterogeneous model architectures across clients~\cite{shao2024selective}.

Among proxy-data approaches, baseline FedMD \cite{FedMD} employs clients to iteratively share per‑sample proxy logits with the server, achieving ML architecture‑agnostic collaboration. The server averages them, and each client distills knowledge from the ensembled logits, giving them an ensemble teacher. Subsequent efforts tried to further reduce the communication overhead and the public‑data requirement rather than abandon it. Communication‑efficient FD \cite{sattler2020communication} compresses the standard FD pipeline by quantising and delta‑coding the proxy logits, and the server performs a second distillation pass, yielding over 100 times bandwidth savings on large models. Distributed distillation \cite{bistritz2020distributed} removes the central server altogether, where clients exchange soft decisions on proxy data over a sparse graph and reach consensus by distributed averaging, reducing communication by two orders of magnitude. FedED~\cite{FedED} method retains a centralized model but employs knowledge distillation to enable local model updates without full parameter sharing. Despite its advantages, FedED does not address challenges associated with non-IID data and heterogeneous model architectures. DS-FL~\cite{DS-FL} uses semi-supervised distillation to boost model size scalability and cope with non-IID data, but ignores heterogeneous client models or handling of scenarios involving unevenly distributed or large-scale data.

Closest to our work is selective-FD~\cite{shao2024selective} that shares selected knowledge between clients and servers by filtering out misleading and ambiguous knowledge, respectively. Selective-FD employs KulSIF-DRE \cite{KulSIF} to filter misleading proxy data and sends only ID proxy data to the server for global aggregation. However, KulSIF-DRE is resource-intensive for clients whose complexity increases exponentially with both dimensionality and number of input data samples \cite{KulSIF, shao2024selective}. Additionally, KuLSIF-DRE requires synthetic \textit{auxiliary data} generated locally on clients. By using auxiliary data, KulSIF-DRE learns the client's private data distribution to distinguish between ID and OOD proxy data based on a predefined threshold. While prior tuning of configurable critical factors (e.g., threshold, auxiliary data characteristics, kernel width parameters, regularization settings, and dataset extrema) increases the performance of the KulSIF-DRE filtering, the computational and memory requirements for the matrix operations significantly influence its real-world applicability. Consequently, implementing Selective-FD~\cite{shao2024selective} on resource-constrained edge devices introduces significant overhead and latency, limiting the practical scalability in federated environments.

\begin{algorithm}[t]
\footnotesize
\caption{EdgeFD using KMeans-DRE algorithm.}
\label{alg:fd_kmeans}
\begin{algorithmic}[1]
\Require server \(\mathcal{S}\), clients \(\mathcal{C}\!\!=\!\!\{1,\dots,C\}\), private data \(\mathcal{D}\!\!=\!\!\{\mathcal{D}_1,\dots,\mathcal{D}_C\}\), private local ML models \(\mathcal{M}\!\!=\!\!\{\mathcal{M}_1,\dots,\mathcal{M}_C\}\), number of training rounds \(R\), ID detection threshold \(T^{ID}\), client proxy data \(\alpha\).
\vspace{0.5em}
\footnotesize
\Procedure{Initialization}{} 
\For {\(c\!\in\!\mathcal{C}\) \textbf{in parallel}}
    \State \(K_{c} \gets \Call{KMeansDRE}{\mathcal{D}_c}\) \Comment{Centroids from private data}
    \State \(\mathcal{M}_c \gets \Call{InitializeLocalModel}{\mathcal{M}_c}\)
    \State \(\mathcal{D}^{\text{proxy}}_{c} \gets \Call{GetProxySubset}{\alpha, \mathcal{D}_{c}}\)
    \State \(\Call{Send}{c, \mathcal{S}, \mathcal{D}^{\text{proxy}}_{c}}\) \Comment{Send proxy subset to the server}
\EndFor
\State \(\mathcal{D}^{\text{proxy}} \gets \Call{ServerAggregateProxy}{\{\mathcal{D}^{\text{proxy}}_c\}_{c\in \mathcal{C}}}\)
\State \(\Call{ShareProxy}{\mathcal{S}, \mathcal{C}, \mathcal{D}^{\text{proxy}}}\) \Comment{Share $\mathcal{D}^{\text{proxy}}$ with clients}
\EndProcedure
\vspace{0.5em}
\Procedure{Training}{}
\For{\(r \gets 1\) \textbf{to} \(R\)}
    \State \(\mathcal{I}_r \gets \Call{SelectRandomIndices}{\mathcal{D}^{\text{proxy}}}\)
    \State \(\{\mathcal{Y}_{c}^{\text{ID}}\}_{c\in \mathcal{C}} \gets \Call{ClientUpdateProxyPredictions}{\mathcal{I}_r}\)
    \vspace{0.1em}
    \State \(\bar{\mathcal{Y}}^{\text{proxy}} \gets \Call{ServerAggregatePredictions}{\{\mathcal{Y}_{c}^{\text{ID}}\}_{c\in \mathcal{C}}}\)
    \State \(\Call{ClientUpdateLocalModel}{\bar{\mathcal{Y}}^{\text{proxy}}}\)
\EndFor
\EndProcedure
\vspace{0.5em}
\Function{ClientUpdateProxyPredictions}{$ \mathcal{I} $}
    \For {\(c\!\in\!\mathcal{C}\) \textbf{in parallel}}
        \State \(\mathcal{D}_{c,r}^{\text{proxy}} \gets \mathcal{D}^{\text{proxy}}[\mathcal{I}]\) \Comment{Proxy subset based on indices $\mathcal{I}$}
        \State \(\mathcal{Y}_{c}^{\text{proxy}} \gets \mathcal{M}_{c}(\mathcal{D}_{c,r}^{\text{proxy}})\) \Comment{Proxy subset predictions}
        \State \(\mathcal{Y}_{c}^{\text{ID}} \gets \textsc{ClientFilter}(\mathcal{D}_c, \mathcal{Y}_{c}^{\text{proxy}}, K_{c}, T^{\text{ID}})\)
        \State \(\Call{Send}{c, \mathcal{S}, \mathcal{Y}_{c}^{\text{ID}}}\)
    \EndFor
    \State \Return $\{\mathcal{Y}_c^{ID}\}_{c\in\mathcal{C}}$
\EndFunction
\vspace{0.5em}
\Function{ClientFilter}{$\mathcal{D}, \mathcal{Y}, K, T^{\text{ID}}$} \Comment{Select only ID samples}
    \State \(\mathcal{Y}^{\text{ID}} \gets \emptyset\)
    \For {\(y\!\in\!\mathcal{Y}\)}
        \State \(x \gets \Call{GetSample}{y}\)
        \If 
        {\(x\!\in\!\mathcal{D}\) \textbf{or} {\(\Call{Distance}{x, K}\!\leq\!T^{\text{ID}}\)}}
            \State \(\mathcal{Y}^{\text{ID}} \gets \mathcal{Y}^{\text{ID}} \cup \{y\}\)
        \EndIf
    \EndFor
    \State \Return \(\mathcal{Y}^{\text{ID}}\)
\EndFunction
\vspace{0.5em}
\Function{ClientUpdateLocalModel}{$ \mathcal{Y}^{\text{proxy}} $}
    \For {\(c\!\in\!\mathcal{C}\) \textbf{in parallel}}
        \State \(\mathcal{M}_{c} \gets \Call{LocalTraining}{\mathcal{M}_{c}, \mathcal{D}_{c}}\) \Comment{Train on private data}
        \State \(\mathcal{M}_{c} \gets \Call{Distill}{\mathcal{M}_{c}, \mathcal{Y}^{\text{proxy}}, \mathcal{D}^{\text{proxy}}}\) \Comment{Knowledge distillation}
    \EndFor
\EndFunction
\end{algorithmic}
\end{algorithm}

\section{Methodology}
\label{sec:methodology}
We propose EdgeFD, a novel method that combines a resource-efficient KMeans model with density-ratio estimation (KMeans-DRE) to reduce resource consumption while eliminating the need for server-side selection. EdgeFD captures the distribution of a client's private data by using KMeans to estimate centroid positions. Then, it estimates the density of the proxy data samples through DRE by calculating the Euclidean distance between each sample and its nearest centroid, discarding those exceeding a predefined threshold as OOD and submitting ID proxy data to the server for global aggregation.

Fig.~\ref{fig:proposed_work} illustrates the global and client side of the proposed EdgeFD method. At the global level, each participating client maintains three key components: private data, a private customized ML model, and a copy of the shared proxy data. Clients generate the proxy data before initializing the FD process by sharing a small percentage of private data, publicly available data, or synthetic data. Since the generation of proxy data lacks a precise definition in FD literature~\cite{qin2024knowledge}, we adopt a similar strategy as other baseline FD methods \cite{shao2024selective, FedMD, FedED, DS-FL} by sharing a small percentage (typically 10\% or 20\%) of the client’s private data. The server, considered a trusted entity, collects a small subset (10-20\%) of each client's private data and creates proxy data distributed back to all participating clients for knowledge distillation. Fig.~\ref{fig:proposed_work} assumes pre-generation of the proxy data.

\begin{table*}[t]
\centering
\caption{Customized ML architectures deployed across EdgeFD clients for MNIST and FashionMNIST datasets.}
\label{tab:mnist-architectures}
\resizebox{\textwidth}{!}{%
\begin{tabular}{@{ }c@{ }c@{ }c@{ }c@{ }c@{ }c@{ }c@{ }c@{ }c@{ }c@{ }}
\toprule
\emph{Client 1}                                                                                                                                                                                                                                                                                                                                                                                                                                                                                                                                       & \emph{Client 2}                                                                                                                                                                                                                                                                                                              & \emph{Client 3}                                                                                                                                                                                                       & \emph{Client 4}                                                                                                                                                                                                                           & \emph{Client 5}                                                                                                                                                                                                                       & \emph{Client 6}                                                                                                                                                                                                                                            & \emph{Client 7}                                                                                                                                                                                                                                             & \emph{Client 8}                                                                                                                                                                                                   & \emph{Client 9}                                                                                                                                                                                                    & \emph{Client 10}                                                                                                                                                                                   \\ \midrule
\begin{tabular}[c]{@{}c@{}}Conv2d(1, 10, 5)\\ Conv2d(10, 20, 5)\\ Linear(320, 50)\\ Linear(50, 10)\end{tabular}                                                                                                                                                                                                                                                                                                                                                                                                                                         & \begin{tabular}[c]{@{}c@{}}Conv2d(1, 16, 3)\\ Conv2d(16, 32, 3)\\ Conv2d(32, 64, 3)\\ Linear(576, 50)\\ Linear(50, 10)\end{tabular}                                                                                                                                                                                            & \multicolumn{1}{l}{\begin{tabular}[c]{@{}l@{}}Conv2d(1, 10, 5)\\ Conv2d(10, 20, 5)\\ Linear(320, 50)\\ Linear(50, 10)\end{tabular}}                                                                                     & \multicolumn{1}{l}{\begin{tabular}[c]{@{}l@{}}Conv2d(1, 12, 3)\\ Conv2d(12, 24, 3)\\ Conv2d(24, 48, 3)\\ Linear(432, 100)\\ Linear(100, 50)\\ Linear(50, 10)\end{tabular}}                                                                  & \multicolumn{1}{l}{\begin{tabular}[c]{@{}l@{}}Conv2d(1, 8, 5)\\ Conv2d(8, 16, 5)\\ Linear(256, 100)\\ Linear(100, 50)\\ Linear(50, 10)\end{tabular}}                                                                                    & \multicolumn{1}{l}{\begin{tabular}[c]{@{}l@{}}Conv2d(1, 6, 7)\\ Conv2d(6, 12, 5)\\ Linear(108, 50)\\ Linear(50, 10)\end{tabular}}                                                                                                                            & \multicolumn{1}{l}{\begin{tabular}[c]{@{}l@{}}Conv2d(1, 32, 3)\\ Conv2d(32, 64, 3)\\ Linear(50176, 50)\\ Linear(50, 10)\end{tabular}}                                                                                                                         & \multicolumn{1}{l}{\begin{tabular}[c]{@{}l@{}}Conv2d(1, 20, 5)\\ Conv2d(20, 30, 5)\\ Linear(480, 50)\\ Linear(50, 10)\end{tabular}}                                                                                 & \multicolumn{1}{l}{\begin{tabular}[c]{@{}l@{}}Conv2d(1, 8, 5)\\ Conv2d(8, 16, 5)\\ Linear(256, 64)\\ Linear(64, 32)\\ Linear(32, 10)\end{tabular}}                                                                   & \multicolumn{1}{l}{\begin{tabular}[c]{@{}l@{}}Conv2d(1, 16, 3)\\ Conv2d(16, 32, 3)\\ Conv2d(32, 64, 3)\\ Linear(64, 100)\\ Linear(100, 10)\end{tabular}}                                             \\ \bottomrule
\end{tabular}}
\end{table*}

\begin{table*}[t]
\centering
\caption{Customized ML architectures deployed across clients in EdgeFD for the CIFAR-10 dataset.}
\label{tab:cifar-architectures}
\resizebox{\textwidth}{!}{%
\begin{tabular}{@{}c@{ }c@{ }c@{ }c@{ }c@{ }c@{ }c@{ }c@{ }c@{ }c@{}}
\toprule
\textit{Client 1} & \textit{Client 2} & \textit{Client 3} & \textit{Client 4} & \textit{Client 5} & \textit{Client 6} & \textit{Client 7} & \textit{Client 8} & \textit{Client 9} & \textit{Client 10} \\
\midrule
\begin{tabular}[c]{@{}c@{}}
\multirow{3}{*}{\(\left[\begin{array}{c}\!\!\text{Conv2d($\cdot$, 64, 3)}\!\!\\[-.1em] \text{BatchNorm2d(64)} \end{array}\right]\)$\times$2}\\ \\ \\
\multirow{3}{*}{\(\left[\begin{array}{c}\!\!\text{Conv2d($\cdot$, 128, 3)}\!\!\\[-.1em] \text{BatchNorm2d(128)} \end{array}\right]\)$\times$2}\\ \\ \\
\multirow{3}{*}{\(\left[\begin{array}{c}\!\!\text{Conv2d($\cdot$, 256, 3)}\!\!\\[-.1em] \text{BatchNorm2d(256)} \end{array}\right]\)$\times$3}\\ \\ \\
Conv2d(128, 256, 1)\\
\multirow{3}{*}{\(\left[\begin{array}{c}\!\!\text{Conv2d($\cdot$, 512, 3)}\!\!\\[-.1em] \text{BatchNorm2d(512)} \end{array}\right]\)$\times$3}\\ \\ \\
Linear(2048, 1024)\\ Linear(1024, 512)\\ Linear(512, 10)\end{tabular}
& \begin{tabular}[c]{@{}c@{}}Conv2d(3, 64, 3)\\ BatchNorm2d(64)\\ Conv2d(64, 128, 3)\\ BatchNorm2d(128)\\ Conv2d(128, 128, 3)\\ BatchNorm2d(128)\\ Conv2d(128, 256, 3)\\ BatchNorm2d(256)\\ Conv2d(128, 256, 1)\\ Conv2d(256, 512, 3)\\ BatchNorm2d(512)\\ Linear(2048, 1024)\\ Linear(1024, 512)\\ Linear(512, 10)\end{tabular}
& \begin{tabular}[c]{@{}c@{}}Conv2d(3, 64, 5)\\ BatchNorm2d(64)\\ Conv2d(64, 128, 5)\\ BatchNorm2d(128)\\ Conv2d(128, 256, 3)\\ BatchNorm2d(256)\\ Linear(16384, 1024)\\ Linear(1024, 512)\\ Linear(512, 10)\end{tabular}
& \begin{tabular}[c]{@{}c@{}}Conv2d(3, 64, 5)\\ BatchNorm2d(64)\\ Conv2d(64, 128, 5)\\ BatchNorm2d(128)\\ Conv2d(128, 256, 3)\\ BatchNorm2d(256)\\ Conv2d(256, 512, 3)\\ BatchNorm2d(512)\\ Linear(32768, 512)\\ Linear(512, 10)\end{tabular}
& \begin{tabular}[c]{@{}c@{}}Conv2d(3, 32, 3)\\ BatchNorm2d(32)\\ Conv2d(32, 64, 3)\\ BatchNorm2d(64)\\ Conv2d(64, 128, 3)\\ BatchNorm2d(128)\\ Conv2d(128, 128, 3)\\ BatchNorm2d(128)\\ Linear(8192, 512)\\ Linear(512, 10)\end{tabular}
& \begin{tabular}[c]{@{}c@{}}Conv2d(3, 32, 3)\\ BatchNorm2d(32)\\ Conv2d(32, 64, 3)\\ BatchNorm2d(64)\\ Conv2d(64, 128, 3)\\ BatchNorm2d(128)\\ Conv2d(64, 128, 1)\\ Conv2d(128, 256, 3)\\ BatchNorm2d(256)\\ Linear(4096, 512)\\ Linear(512, 10)\end{tabular}
& \begin{tabular}[c]{@{}c@{}}Conv2d(3, 32, 5)\\ BatchNorm2d(32)\\ Conv2d(32, 64, 5)\\ BatchNorm2d(64)\\ Conv2d(64, 128, 3)\\ BatchNorm2d(128)\\ Conv2d(128, 256, 3)\\ BatchNorm2d(256)\\ Linear(16384, 1024)\\ Linear(1024, 256)\\ Linear(256, 10)\end{tabular}
& \begin{tabular}[c]{@{}c@{}}Conv2d(3, 32, 3)\\ BatchNorm2d(32)\\ Conv2d(32, 64, 3)\\ BatchNorm2d(64)\\ Conv2d(64, 128, 3)\\ BatchNorm2d(128)\\ Conv2d(64, 128, 1)\\ Linear(8192, 512)\\ Linear(512, 10)\end{tabular}
& \begin{tabular}[c]{@{}c@{}}Conv2d(3, 64, 3)\\ BatchNorm2d(64)\\ Conv2d(64, 128, 3)\\ BatchNorm2d(128)\\ Conv2d(128, 128, 3)\\ BatchNorm2d(128)\\ Linear(8192, 512)\\ Linear(512, 256)\\ Linear(256, 10)\end{tabular}
& \begin{tabular}[c]{@{}c@{}}Conv2d(3, 64, 3)\\ BatchNorm2d(64)\\ Conv2d(64, 128, 3)\\ BatchNorm2d(128)\\ Conv2d(128, 256, 3)\\ BatchNorm2d(256)\\ Linear(16384, 1024)\\ Linear(1024, 10)\end{tabular} \\ \bottomrule
\end{tabular}}
\end{table*}

\subsection{EdgeFD Workflow and Algorithm}
\label{subsec:edgefd_workflow_and_algo}
The general EdgeFD workflow, summarized in Algorithm~\ref{alg:fd_kmeans}, consists of two phases. 

\subsubsection{Initialization phase (lines 1--10)} Each client \(c\) trains a KMeans model on its private dataset \(\mathcal{D}_c\) to compute centroid positions \(K_c\) and initializes its local model \(\mathcal{M}_c\). Next, each client selects a fraction \(\alpha\) of its private data to form a proxy subset \(\mathcal{D}_c^{\text{proxy}}\) and sends this subset to the server. The server aggregates these proxy subsets into a shared proxy dataset \(\mathcal{D}^{\text{proxy}}\) and redistributes it to all clients.

\subsubsection{Training phase (lines 11--18)} For each training round~\(r\), the server and clients perform the following operations:
\begin{itemize}
    \item Server randomly selects a set of proxy data indices \(\mathcal{I}_r\) from \(\mathcal{D}^{\text{proxy}}\) and sends them all clients (lines 13--14);
    \item Each client performs the following operations:
    \begin{itemize}
        \item Extracts the proxy samples for \(\mathcal{I}_r\) and computes predictions using its local model (lines 21--22);
        \item Applies the \textsc{ClientFilter} function (line 23) to select only ID predictions \(\mathcal{Y}_c^{\text{ID}}\) for which (i) the corresponding sample is present in the private dataset \(\mathcal{D}_c\) or (ii) its Euclidean distance to the centroids \(K_c\) is below the threshold \(T^{\text{ID}}\) (lines 28--37);
        \item Sends \(\mathcal{Y}_c^{\text{ID}}\) to the server (line 24);
    \end{itemize}
    \item Server aggregates the ID predictions from all clients to calculate an averaged proxy prediction \(\bar{\mathcal{Y}}^{\text{proxy}}\) and distributes it back (lines 15--16);
    \item Each client updates its local model \(\mathcal{M}_c\) by training on the private data \(\mathcal{D}_c\) and performing knowledge distillation using the received \(\bar{\mathcal{Y}}^{\text{proxy}}\) (lines 40--41);
    \item The algorithm repeats this training cycle for \(R\) rounds.
\end{itemize}

Client-side filtering method \textsc{ClientFilter} (lines 28--37) is essential for EdgeFD, particularly without a pre-trained teacher model on the server. Client models \(\mathcal{M}_c\) trained on private data typically generate unreliable predictions for proxy data from other clients~\cite{shao2024selective}, as these samples are often OOD relative to the local data. EdgeFD implements robust client-side filtering using KMeans-DRE to prevent negative knowledge transfer and evaluates each proxy sample from other clients against the client's local data distribution, classifying them as either ID or OOD. By transmitting only predictions from ID proxy samples to the server, this filtering preserves the effectiveness of the knowledge distillation process.

\section{Experimental Setup and Evaluation}
\label{sec:experiments}
To evaluate EdgeFD, we conduct experiments on three benchmark image classification datasets: MNIST~\cite{mnist}, FashionMNIST~\cite{fashionmnist}, and CIFAR10~\cite{cifar10}. Each dataset contains 10 classes with 50,000 training and 10,000 test samples. Our evaluation simulates real-world heterogeneous scenarios by implementing varying class distributions across clients (see Sec.~\ref{subsec:datadistrib}) and the corresponding strategy for classifying ID and OOD proxy samples (see Sec.~\ref{subsec:proxydatacls}). The experimental setup involves ten clients, each running a distinct ML model architecture described in Table \ref{tab:mnist-architectures} and Table \ref{tab:cifar-architectures}.

\subsection{Data Distribution Scenarios}
\label{subsec:datadistrib}
We implement three data distribution scenarios to evaluate EdgeFD under realistic conditions.
\paragraph{Strong Non-IID} Each client possesses data samples from a unique subset of output labels that do not overlap with other clients. The KMeans model initialized with a single centroid captures the distinct data distribution pattern specific to each client's local dataset, reflecting real-world scenarios with edge devices collecting specialized data based on their deployment context.
\paragraph{Weak Non-IID} Each client has samples from multiple overlapping output labels, simulated by randomly assigning three output labels to each client from the available ten labels. The KMeans model initializes one centroid per output label on each client to effectively capture the data distribution.
\paragraph{IID} We implement an IID baseline scenario with data samples uniformly distributed across all clients.

\subsection{Proxy Data Classification Strategy}
\label{subsec:proxydatacls}
The KMeans model captures the private data distribution and learns the centroid positions. In the strong non-IID case, KMeans-DRE computes the Euclidean distance between the single centroid position and the proxy samples, classifying them as ID or OOD using client-specific predefined thresholds.

The weak non-IID and IID scenarios require a different approach due to multiple overlapping labels per client. Here, KMeans-DRE evaluates distances between each proxy sample and multiple centroid positions. A proxy sample is ID if it belongs to at least one cluster, and OOD otherwise.

\begin{table}[t]
\centering
\renewcommand{\arraystretch}{0.75}
\caption{Comparative evaluation of EdgeFD with related methods on three datasets and IID conditions, highlighting the best method in bold and the second-best underlined.}
\label{tab:results}
\resizebox{\columnwidth}{!}{%
\begin{tabular}{@{}lccc@{}}
\toprule
\emph{Strong Non-IID} & \emph{MNIST} & \emph{FashionMNIST} & \emph{CIFAR10} \\ \midrule
IndLearn                & 10.00          & 10.00                 & 10.00            \\ \midrule
FedMD~\cite{FedMD}      & 88.71          & 64.63                 & 15.78            \\
FedED~\cite{FedED}      & 11.92          & 12.52                 & 12.04            \\
DS-FL~\cite{DS-FL}      & 35.25          & 35.98                 & 12.07            \\
FKD~\cite{FKD}          & 10.00          & 10.00                 & 10.00            \\
PLS~\cite{PLS}          & 10.00          & 10.00                 & 10.00            \\
\vspace{1px}
Selective-FD~\cite{shao2024selective}    & \underline{94.68}          & \underline{75.31}                 & \underline{80.98}            \\
EdgeFD           & \textbf{98.92} & \textbf{88.55}        & \textbf{82.57}   \\
\midrule
\emph{Weak Non-IID}   & \emph{MNIST} & \emph{FashionMNIST} & \emph{CIFAR10} \\ \midrule
IndLearn                & 19.96          & 19.82                 & 19.52            \\ \midrule
FedMD~\cite{FedMD}      & 95.16          & 74.83                 & 84.31            \\
FedED~\cite{FedED}      & 60.26          & 37.12                 & 56.13            \\
DS-FL~\cite{DS-FL}      & 47.87          & 39.22                 & 52.51            \\
FKD~\cite{FKD}          & 19.98          & 19.71                 & 19.51            \\
PLS~\cite{PLS}          & 19.97          & 19.70                 & 19.52            \\
\vspace{1px}
Selective-FD~\cite{shao2024selective}    & \underline{96.30}   & \underline{77.27}       & \textbf{85.38}           \\
EdgeFD           & \textbf{98.88} & \textbf{88.74}        & \underline{84.88}   \\ \midrule
\emph{IID}            & \emph{MNIST} & \emph{FashionMNIST} & \emph{CIFAR10} \\ \midrule
IndLearn                & 98.83          & 88.80                 & 85.53            \\ \midrule
\vspace{1px}
FedMD~\cite{FedMD}      & \underline{98.63}          & \underline{87.25}                 & 86.31            \\
FedED~\cite{FedED}      & 98.26          & 86.88                 & \textbf{86.87}            \\
DS-FL~\cite{DS-FL}      & 98.56          & 86.62                 & 85.82            \\
FKD~\cite{FKD}          & 98.44          & 86.14                 & 84.10            \\
PLS~\cite{PLS}          & 98.48          & 86.52                 & 84.77            \\
Selective-FD~\cite{shao2024selective}    & 98.60     & 87.16     & 86.06            \\
EdgeFD           & \textbf{99.08} & \textbf{89.90}        & \underline{86.37}   \\ \bottomrule
               & \multicolumn{1}{l}{} & \multicolumn{1}{l}{} & \multicolumn{1}{l}{}
\end{tabular}
}
\end{table}

\subsection{State-of-the-Art Comparison}
\label{subsec:performanceeval}
We evaluate the EdgeFD accuracy based on KMeans-DRE, as discussed in Sec.~\ref{subsec:edgefd_workflow_and_algo}, against six SoTA FD methods, keeping the original reported results without confounding fair performance comparison to avoid methodological drifts:
\paragraph{FedMD\cite{FedMD}} is the earliest FD method;
\paragraph{FedED~\cite{FedED} and DS‑FL~\cite{DS-FL}} are two other FD methods that employ proxy data to transfer knowledge;
\paragraph{FKD~\cite{FKD} and PLS~\cite{PLS}} are data-free FD methods that share class-wise average predictions among clients;
\paragraph{Selective‑FD~\cite{shao2024selective}} closest to our work, utilizes DRE techniques, and sends ID proxy logits to the server by employing KuLSIF‑DRE~\cite{KulSIF}.

\subsection{Evaluation Results}
Table~\ref{tab:results} presents the evaluation results covering strong and weak non-IID, and IID scenarios, where  IndLearn refers to the average test accuracy with independently trained clients. 

\paragraph{Strong non-IID} Baseline methods exhibit significant performance limitations. While FedMD achieves 88.57\%  accuracy on MNIST, its performance degrades substantially to 64.63\% on FashionMNIST, and 15.78\% on CIFAR10, indicating decreased effectiveness with complex datasets. FedED, FKD, and PLS achieve approximately 10\% accuracy, indicating their inability to handle strong non-IID cases. In contrast, EdgeFD achieves 98.92\% accuracy on MNIST, 88.55\% on FashionMNIST, and 82.57\% on CIFAR10, exceeding all baseline methods, including the best-performing Selective-FD. The EdgeFD advantage is due to the client-side filtering, which effectively manages extreme data heterogeneity.

\paragraph{Weak non-IID} While existing methods show improved performance compared to strong non-IID cases, they still face limitations. FedED and DS-FL achieve moderate accuracy levels of 60.26\% and 47.87\% on MNIST, respectively, and relatively lower performance on FashionMNIST and CIFAR10. EdgeFD maintains consistent high performance, achieving 98.88\% on MNIST, 88.74\% on FashionMNIST, and 84.88\% on CIFAR10, demonstrating effective handling of class overlap in non-IID conditions.

\paragraph{IID} With uniform data distribution across clients, most methods achieve relatively high accuracy. Traditional FD approaches such as FedMD and FedED perform well, achieving over 98\% on MNIST, while methods like FKD and PLS show slightly lower performance. The consistent high performance of EdgeFD showcases its particular strength in challenging non-IID environments while maintaining competitive performance in IID cases.

\section{Discussion and Analysis}
\label{sec:discussion_and_analysis}
We discuss four performance aspects of EdgeFD: computational complexity and resource consumption of KMeans-DRE compared to KuLSIF-DRE~\cite{KulSIF} in FD methods~\cite{shao2024selective}, critical performance factors, privacy threats in feature-based FD methods, and limitations highlighting future research directions.

\begin{figure}[t]
    \centering
    \begin{subfigure}[b]{0.48\columnwidth}
        \centering
        \includegraphics[width=\columnwidth]{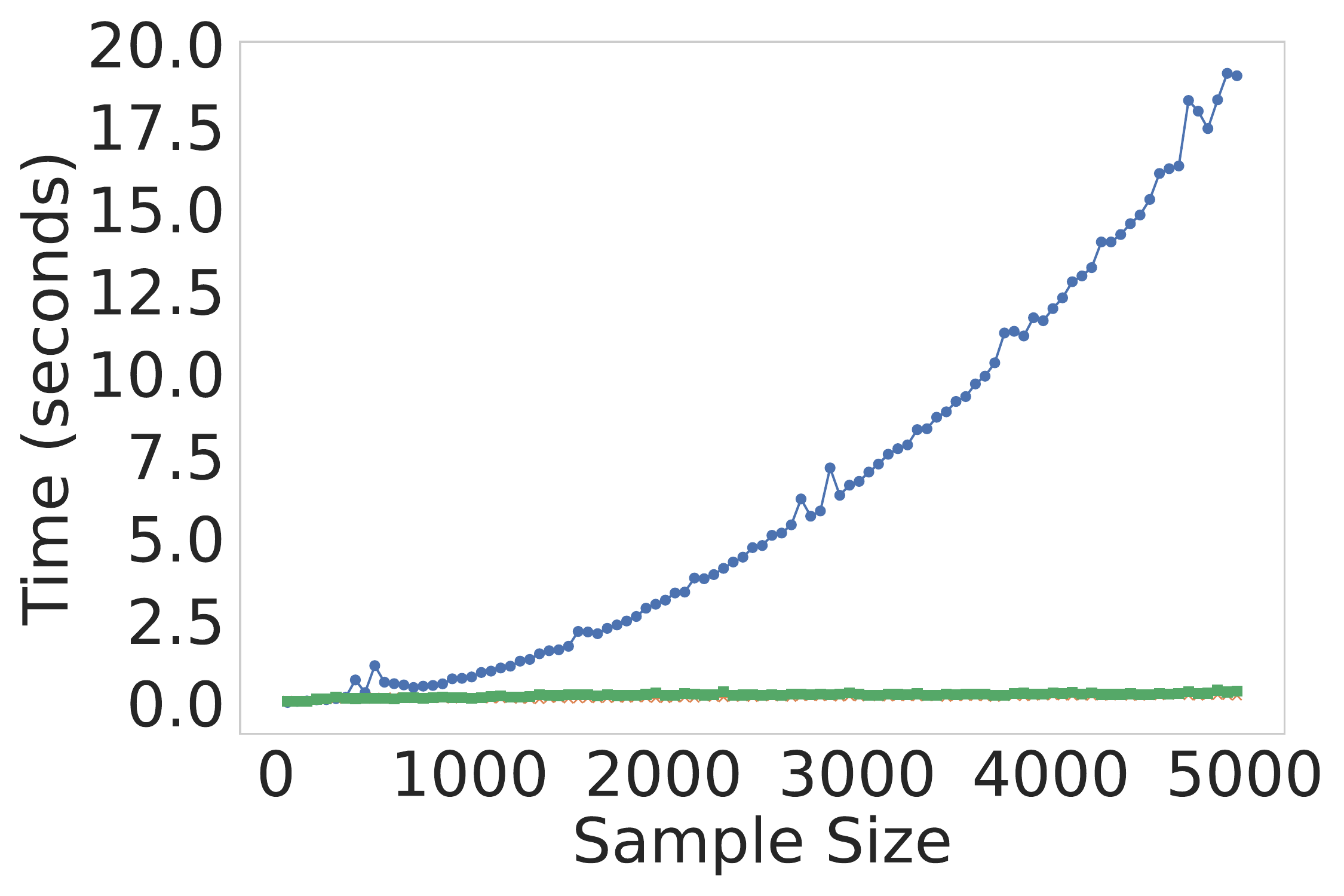}
        \caption{Learning time.}
        \label{fig:training_time_complexity}
    \end{subfigure}
    \hfill
    \begin{subfigure}[b]{0.48\columnwidth}
        \centering
        \includegraphics[width=\columnwidth]{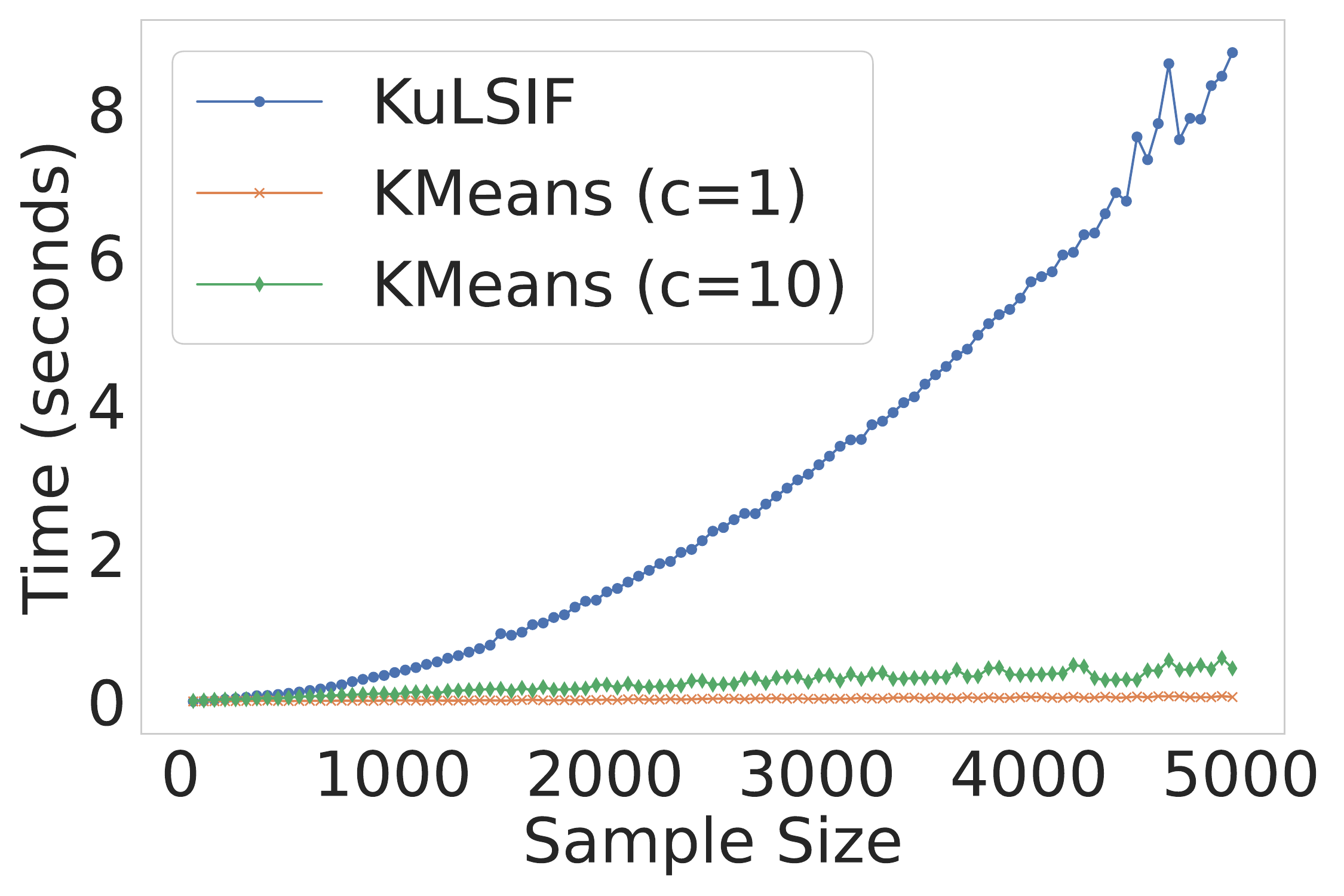}
        \caption{Estimation time.}
        \label{fig:estimation_time_complexity}
    \end{subfigure}
    \hfill
    \begin{subfigure}[b]{0.48\columnwidth}
        \centering
        \includegraphics[width=\columnwidth]{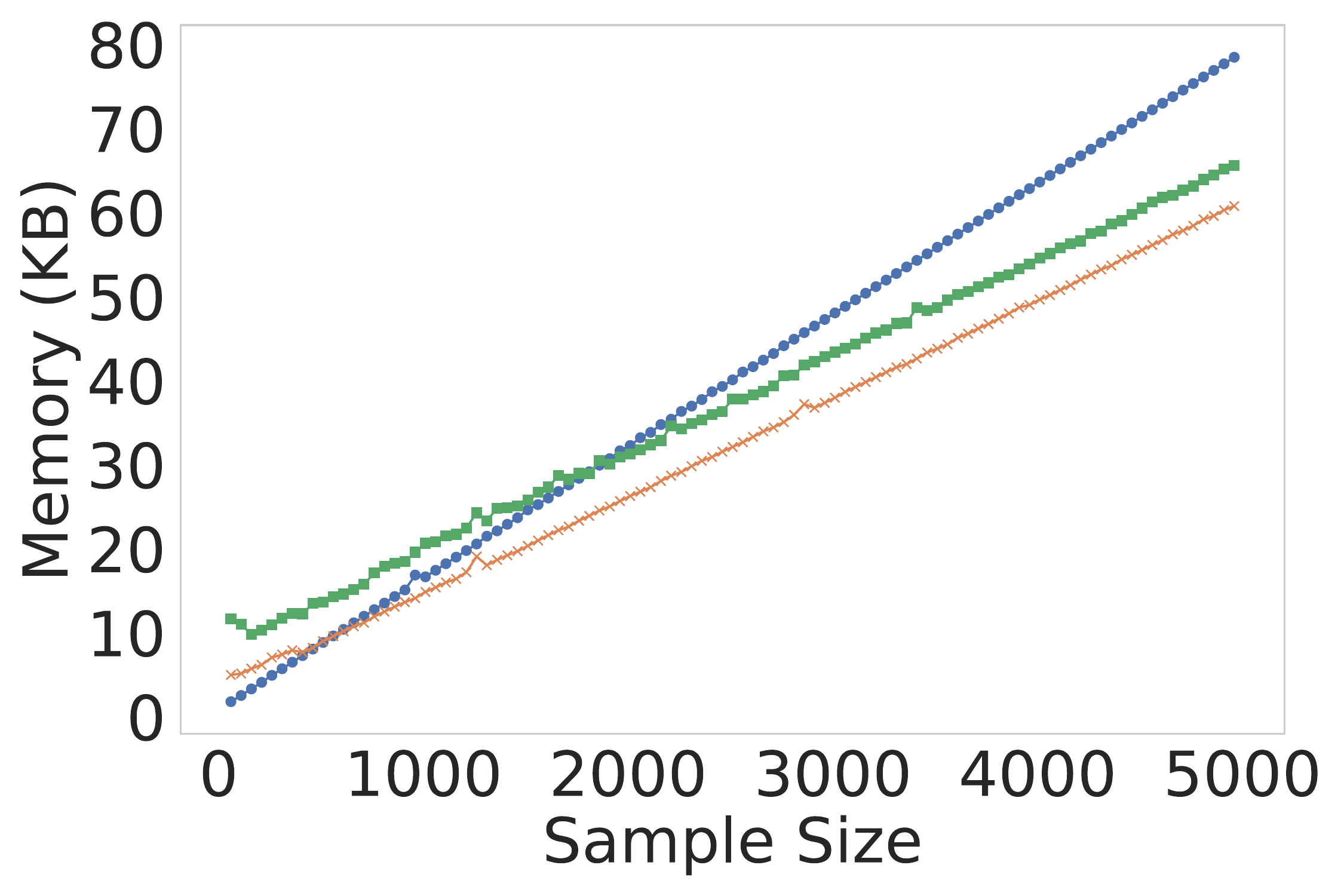}
        \caption{Learning memory.}
        \label{fig:train_memory_complexity}
    \end{subfigure}
    \hfill
    \begin{subfigure}[b]{0.48\columnwidth}
        \centering
        \includegraphics[width=\columnwidth]{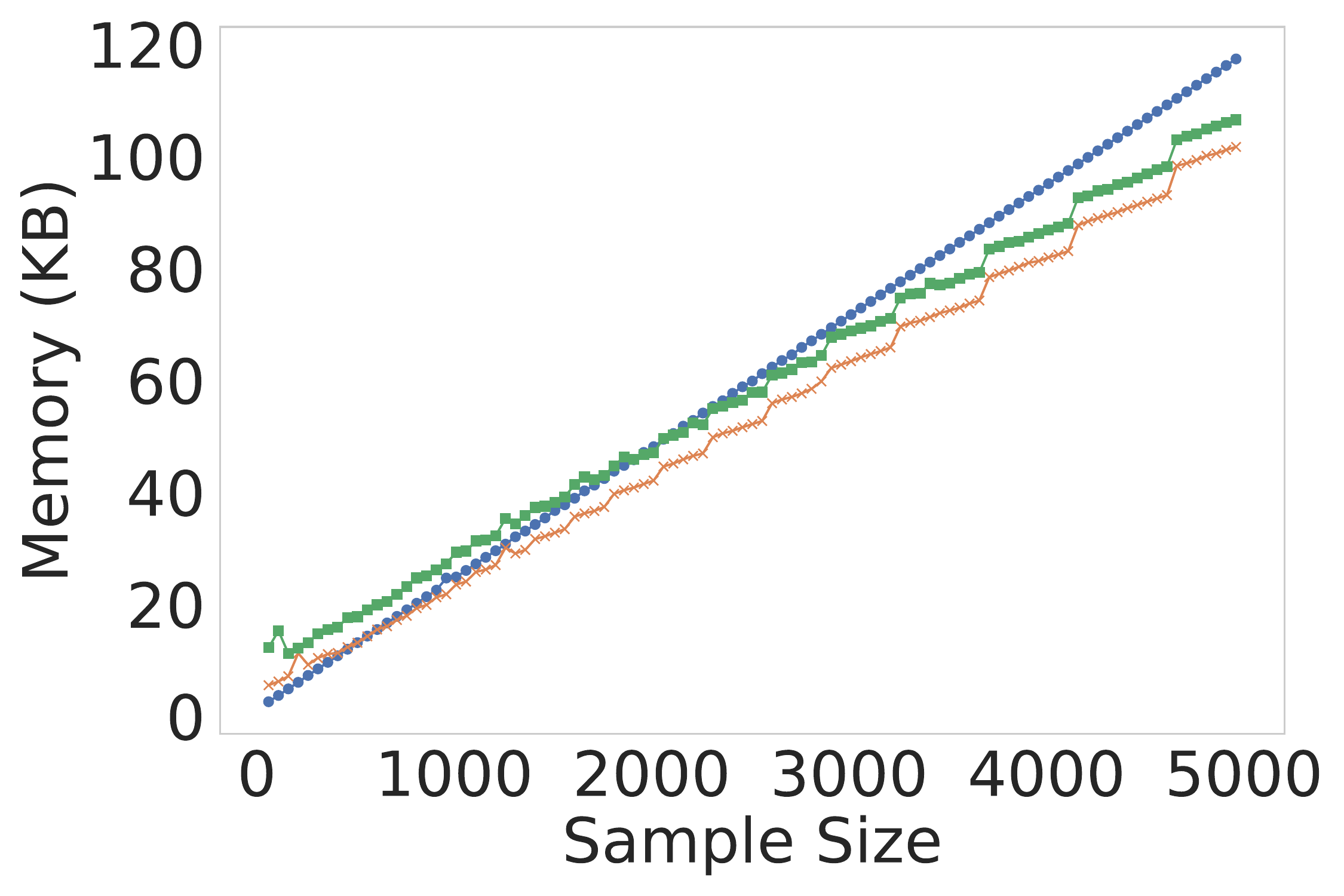}
        \caption{Estimation memory.}
        \label{fig:estimation_memory_complexity}
    \end{subfigure}
    \caption{Learning and estimation time and memory comparison between KuLSIF-DRE~\cite{KulSIF} and KMeans-DRE (with 1 and 10 centroids) for 50-dimensional data.}
    \label{fig:complexity}
\end{figure}

\subsection{Resource Consumption}
\begin{figure}[t]
     \centering
     \begin{subfigure}[b]{.75\columnwidth}
         \centering
         \includegraphics[width=\columnwidth]{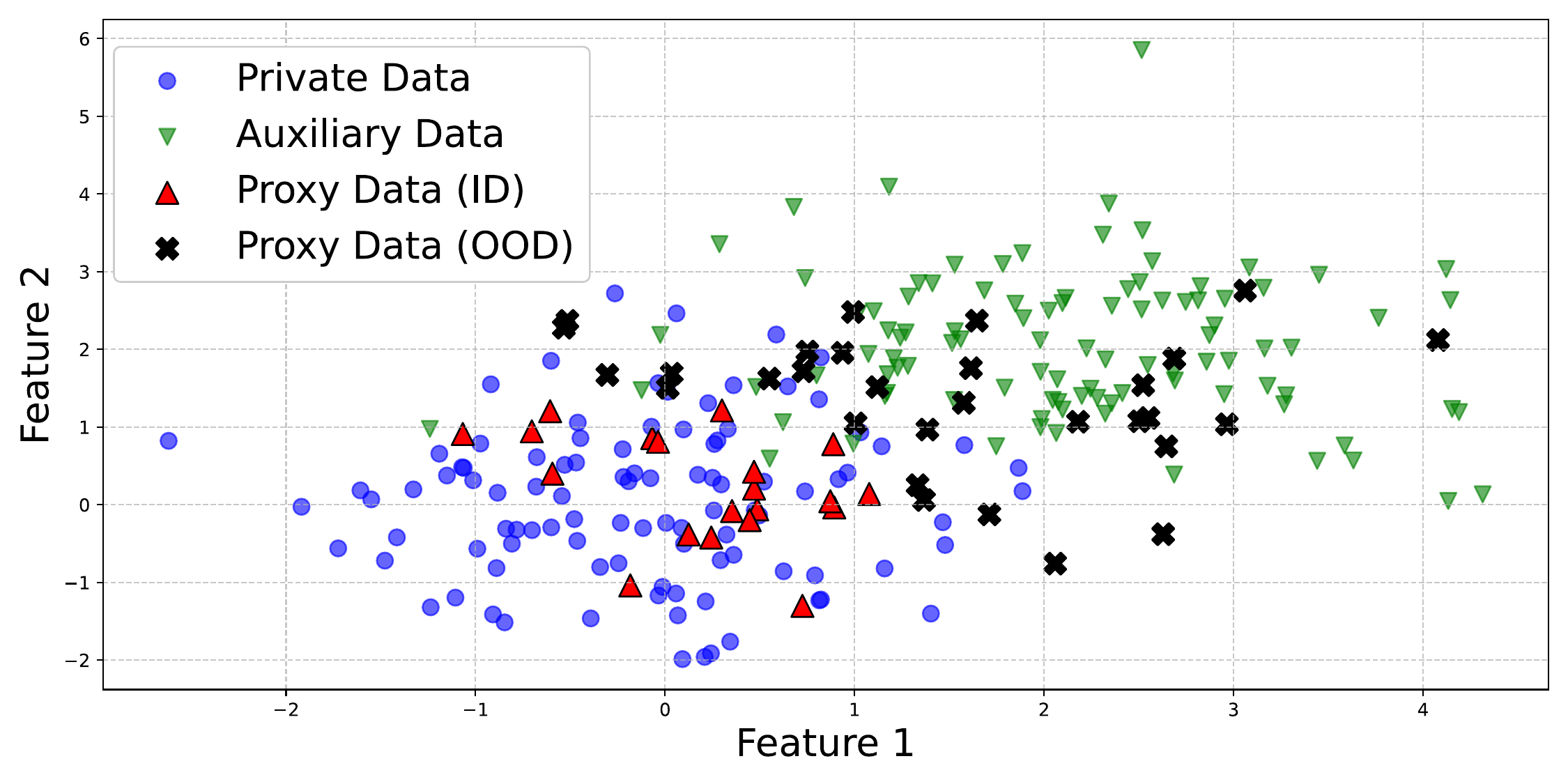}
         \caption{KuLSIF-DRE \cite{KulSIF}.}
         \label{fig:kulsif}
     \end{subfigure}
     \vfill
     \begin{subfigure}[b]{.75\columnwidth}
         \centering
         \includegraphics[width=\columnwidth]{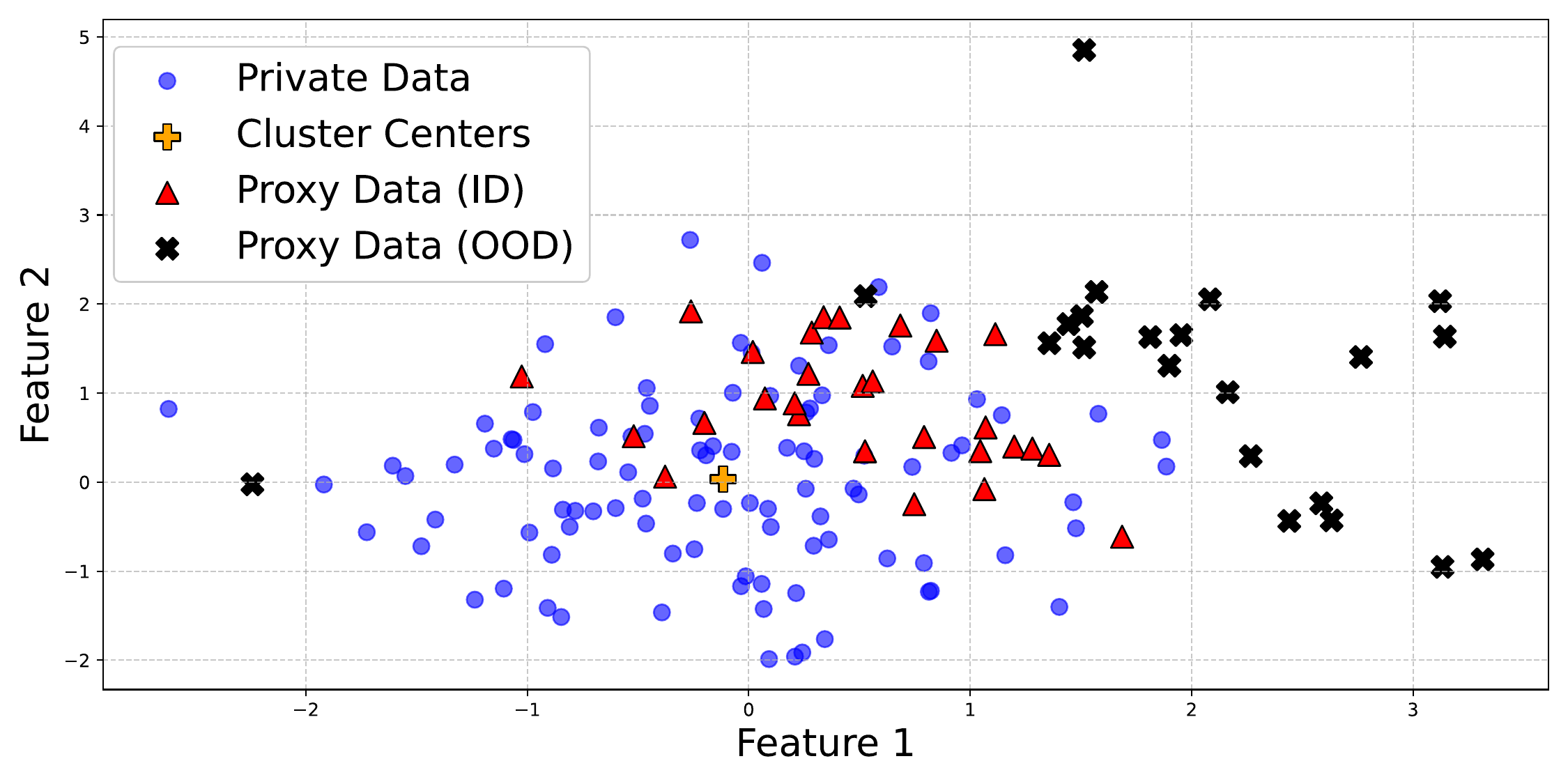}
         \caption{KMeans-DRE.}
         \label{fig:kmeans}
     \end{subfigure}
     \caption{Density ratio estimation comparison using randomly sampled two-feature data.
     }
     \label{fig:kulsif_vs_kmeans}
\end{figure}

A DRE operates in two phases: \textit{learn}, which captures distribution density from private data, and \textit{estimate}, which evaluates test proxy data density using the learned distribution. Fig.~\ref{fig:complexity} presents a comparative analysis of time and space complexity across both phases for KuLSIF-DRE and KMeans-DRE implementations. The analysis reveals distinct differences in computational efficiency, further explored in Appendix~\ref{appendix:bigO}. KuLSIF-DRE exhibits exponential time complexity growth with increasing sample size for both phases. In contrast, KMeans-DRE demonstrates linear time complexity with minimal slope across both phases. While memory complexity grows linearly with the sample size of both DREs, KMeans-DRE reduces memory consumption at large sample sizes. This advantage derives from three key design features:
\begin{enumerate}
    \item A client-side filter that learns private data distribution without auxiliary data required by statistical-DRE methods illustrated in Fig.~\ref{fig:kulsif_vs_kmeans}, enabling lightweight operation;
    \item KMeans model integration that captures data distribution and determines centroid positions without large matrices and parameters as in statistical DREs (see Sec.~\ref{sec:related_works});
    \item An optimized two-stage client-side filtering process that eliminates redundant computation in proxy data evaluation for ID and OOD classification.
\end{enumerate}

\subsection{Time and Space Complexity}
\label{appendix:bigO}
To compare KuLSIF-DRE~\cite{KulSIF} and KMeans-DRE, we compute the time and space complexity of both density-ratio estimation approaches (see Table~\ref{tab:complexity}).

\subsubsection{KuLSIF-DRE} The key parameters are auxiliary samples $m$, private samples $n$, test samples $t$, and dimensionality of the sample data $d$. 

\subsubsection{KuLSIF-DRE learning} involves three components:
\paragraph{\(K_{11}\in\mathbb{R}^{m\times m}\)} matrix computes pairwise Euclidean distances between auxiliary samples followed by a Gaussian kernel, which leads to a time complexity of \(\mathcal{O}\left(m^2\cdot d\right)\) and a space complexity of \(\mathcal{O}\left(m^2\right)\).
\paragraph{\(K_{12}\in\mathbb{R}^{n\times m}\)} matrix computes the distances between private samples and auxiliary samples, with time complexity \(\mathcal{O}(n\cdot m\cdot d)\), and space complexity \(\mathcal{O}(n\cdot m)\).
\paragraph{\(\alpha\)-vector\(\,\in\!\mathbb{R}^{m}\)} derived through a series of matrix operations, including an inversion of the modified \(K_{11}\) and matrix-vector multiplications has a time complexity of \(\mathcal{O}(m^{3})\) and a space complexity of \(\mathcal{O}(m^{2})\).
\subsubsection{KuLSIF-DRE estimation} computes the Euclidean distance between each test sample and both private and auxiliary samples, with \(\mathcal{O}(t\cdot n\cdot d)\) time and \(\mathcal{O}(t\cdot m\cdot d)\) space complexity. Distance calculations dominate other operations, including matrix exponentiation and dot products, leading to an overall time complexity of \(\mathcal{O}(t\cdot(n+m)\cdot d)\). Storage requirements for the distance matrices dominate the space complexity, resulting in a combined space complexity of \(\mathcal{O}(t\cdot(n+m))\).
\subsubsection{KMeans-DRE learning} involves fitting the KMeans clustering algorithm and storing its outputs. The time complexity is \(\mathcal{O}(k\cdot n\cdot c\cdot d)\), where \(n\) is the number of private samples, \(c\) the number of clusters, \(d\) the dimensionality of each sample, and \(k\) the number of convergence iterations. For space complexity, the centroids and labels  require \(\mathcal{O}(c\cdot d + n)\) space.
\subsubsection{KMeans-DRE estimation} computational complexity involves evaluating of cluster membership with an Euclidean distance to each \(c\) cluster centroids for each test sample, an operation with a time complexity of \(\mathcal{O}(t\cdot c\cdot d)\) and a space complexity  of \(\mathcal{O}(c\cdot d + t)\), due to the storage of the boolean results from the threshold for each sample. 

Table~\ref{tab:complexity} indicates that learning and estimation in KuLSIF-DRE are significantly resource-intensive, especially for large datasets and high-dimensional data, questioning their scalability in practical applications.

\begin{table}[t]
\centering
\caption{KuLSIF-DRE versus KMeans-DRE time and space complexity comparison.}
\label{tab:complexity}
\resizebox{\columnwidth}{!}{%
\begin{tabular}{@{}l@{}cc@{}}
\toprule
\textit{KuLSIF-DRE} & \textit{Time}  & \textit{Space} \\ \midrule
$K_{11}$       & $\mathcal{O}\left(m^2\cdot d\right)$ & $\mathcal{O}\left(m^2\right)$ \\ 
$K_{12}$       & $\mathcal{O}(n\cdot m\cdot d)$ & $\mathcal{O}(n\cdot m)$ \\
$\alpha$-vector& $\mathcal{O}(m^3)$ & $\mathcal{O}\left(m^2\right)$ \\
Learning       & $\mathcal{O}\left(m^3 + m^2\cdot d + n \cdot m \cdot d\right)$ & $\mathcal{O}\left(m^2 + n\cdot m\right)$ \\
Estimation  & $\mathcal{O}(t\cdot(n+m)\cdot d)$ & $\mathcal{O}(t\cdot(n+m))$ \\ \midrule
\textit{KMeans-DRE} & \textit{Time}  & \textit{Space} \\ \midrule
Learning       & $\mathcal{O}(k \cdot n\cdot c\cdot d)$ & $\mathcal{O}(c\!\cdot\!d + n)$ \\
Estimation       & $\mathcal{O}(t\cdot c \cdot d)$ & $\mathcal{O}(c\!\cdot\!d + t)$ \\ \bottomrule
\end{tabular}}
\end{table}

\subsection{Key Performance Factors}
We addressed three performance factors concerning data distribution, threshold impact on accuracy, and proxy data for the MNIST, FashionMNIST, and CIFAR-10 datasets, where CIFAR-10* represents features extracted using pre-trained ResNet18 on ImageNet. Table \ref{tab:cifar-architectures} lists the ML architectures for CIFAR-10 dataset.

\begin{figure}[t]
    \centering
    \begin{subfigure}[b]{0.24\textwidth}
        \centering
        \includegraphics[width=\columnwidth]{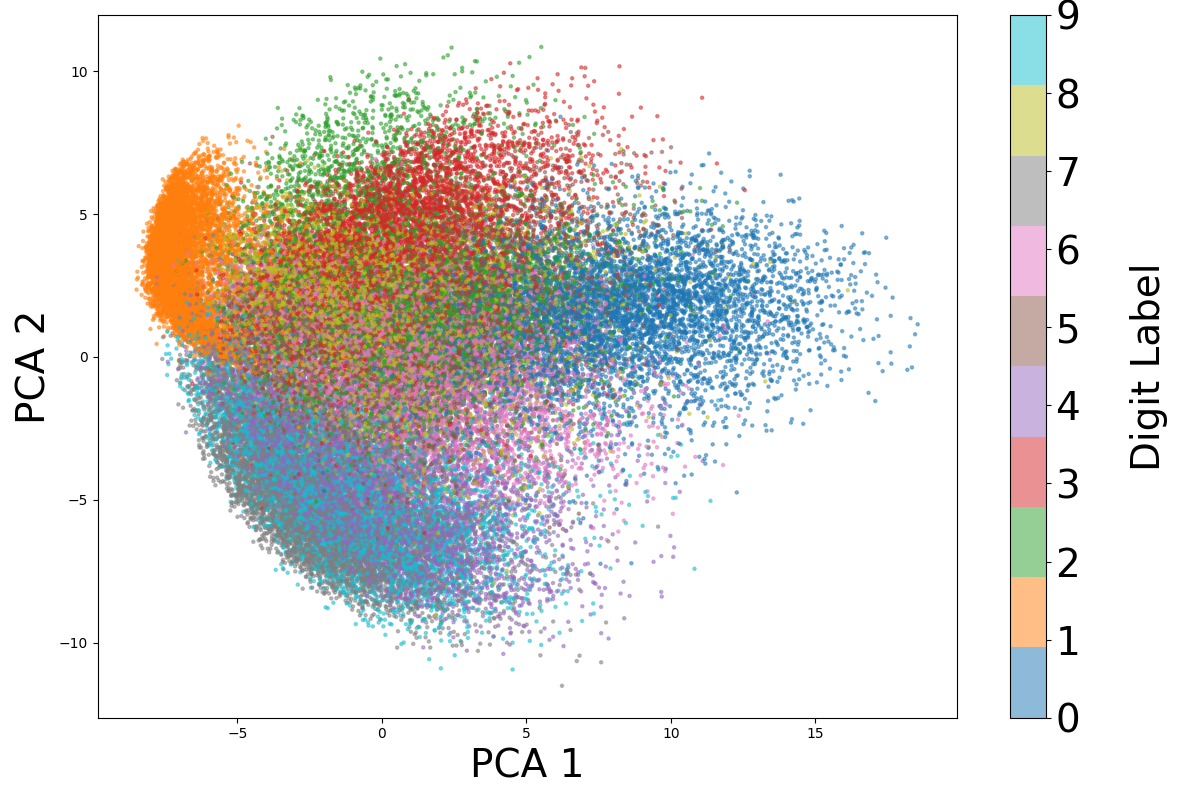}
        \caption{MNIST.}
        \label{fig:mnist}
    \end{subfigure}
    \hfill
    \begin{subfigure}[b]{0.24\textwidth}
        \centering
        \includegraphics[width=\columnwidth]{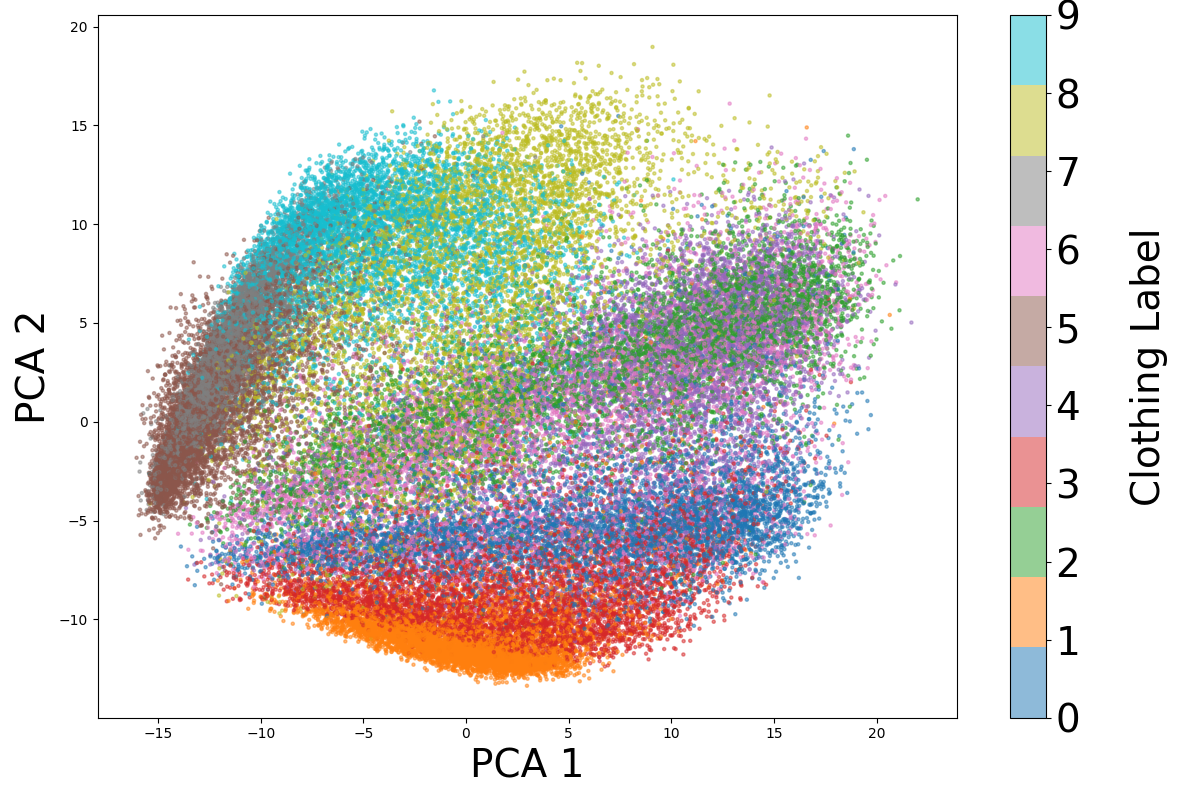}
        \caption{FashionMNIST.}
        \label{fig:fashionmnist}
    \end{subfigure}
    \hfill
    \begin{subfigure}[b]{0.24\textwidth}
        \centering
        \includegraphics[width=\columnwidth]{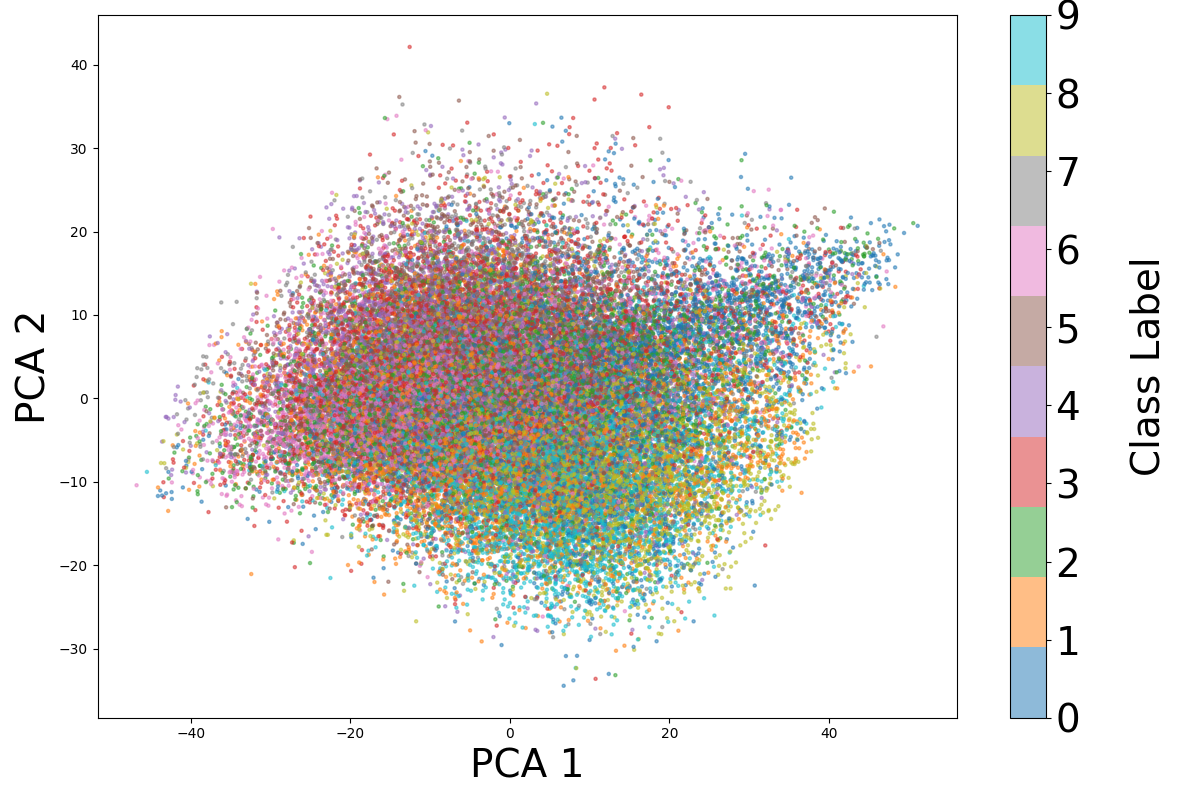}
        \caption{CIFAR-10.}
        \label{fig:cifar10}
    \end{subfigure}
    \hfill
    \begin{subfigure}[b]{0.24\textwidth}
        \centering
        \includegraphics[width=\columnwidth]{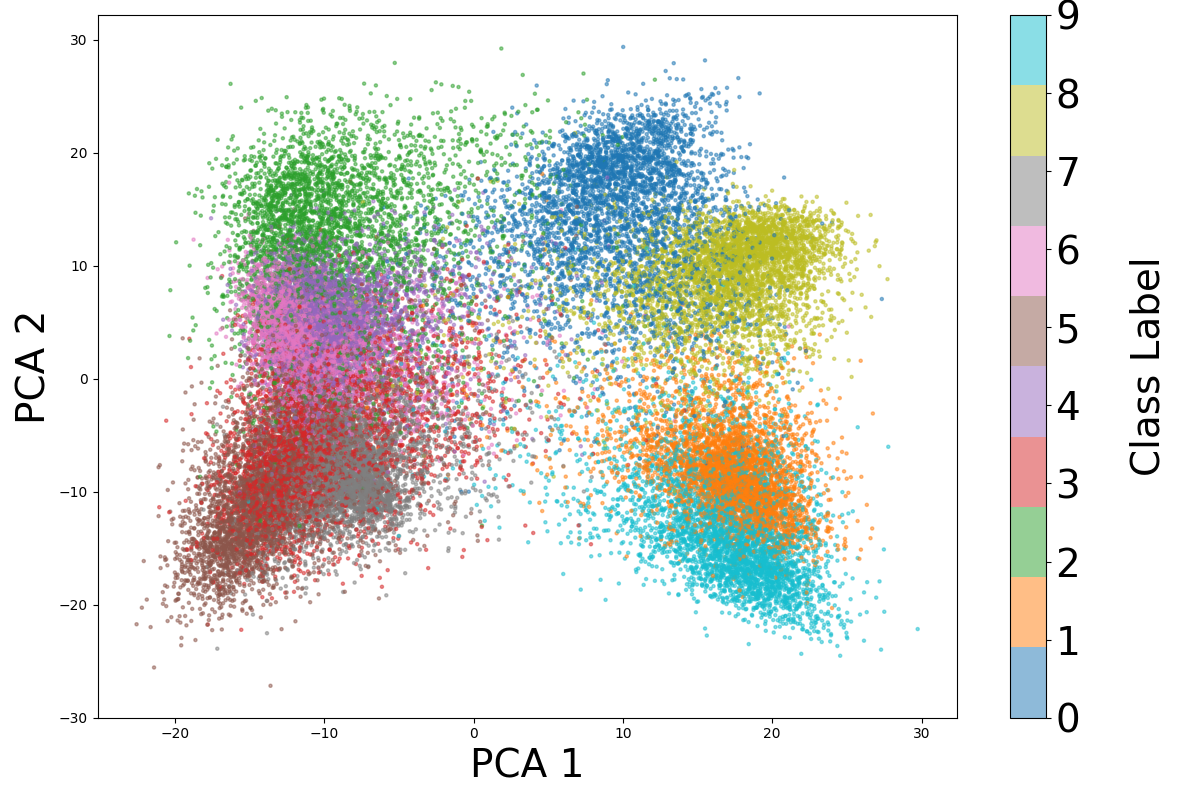}
        \caption{CIFAR10*.}
        \label{fig:cifar10_clusters}
    \end{subfigure}
    \caption{Principal component analysis of various datasets.}
\end{figure}
    
\begin{figure*}[t]
    \centering
    \begin{subfigure}[b]{0.32\textwidth}
        \centering
        \includegraphics[width=\columnwidth]{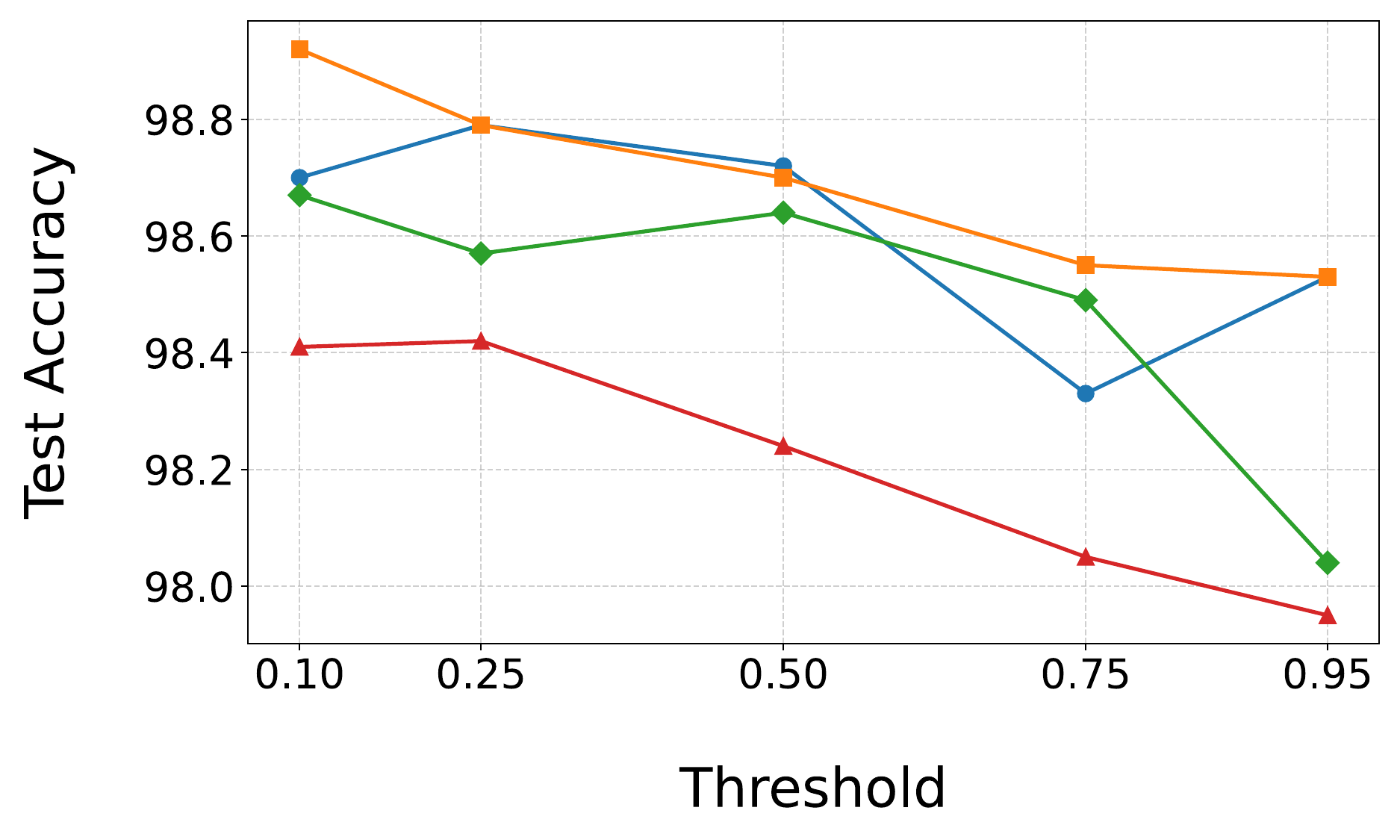}
        \caption{MNIST.}
        \label{fig:mnist_proxy}
    \end{subfigure}
    \hfill
    \begin{subfigure}[b]{0.32\textwidth}
        \centering
        \includegraphics[width=\columnwidth]{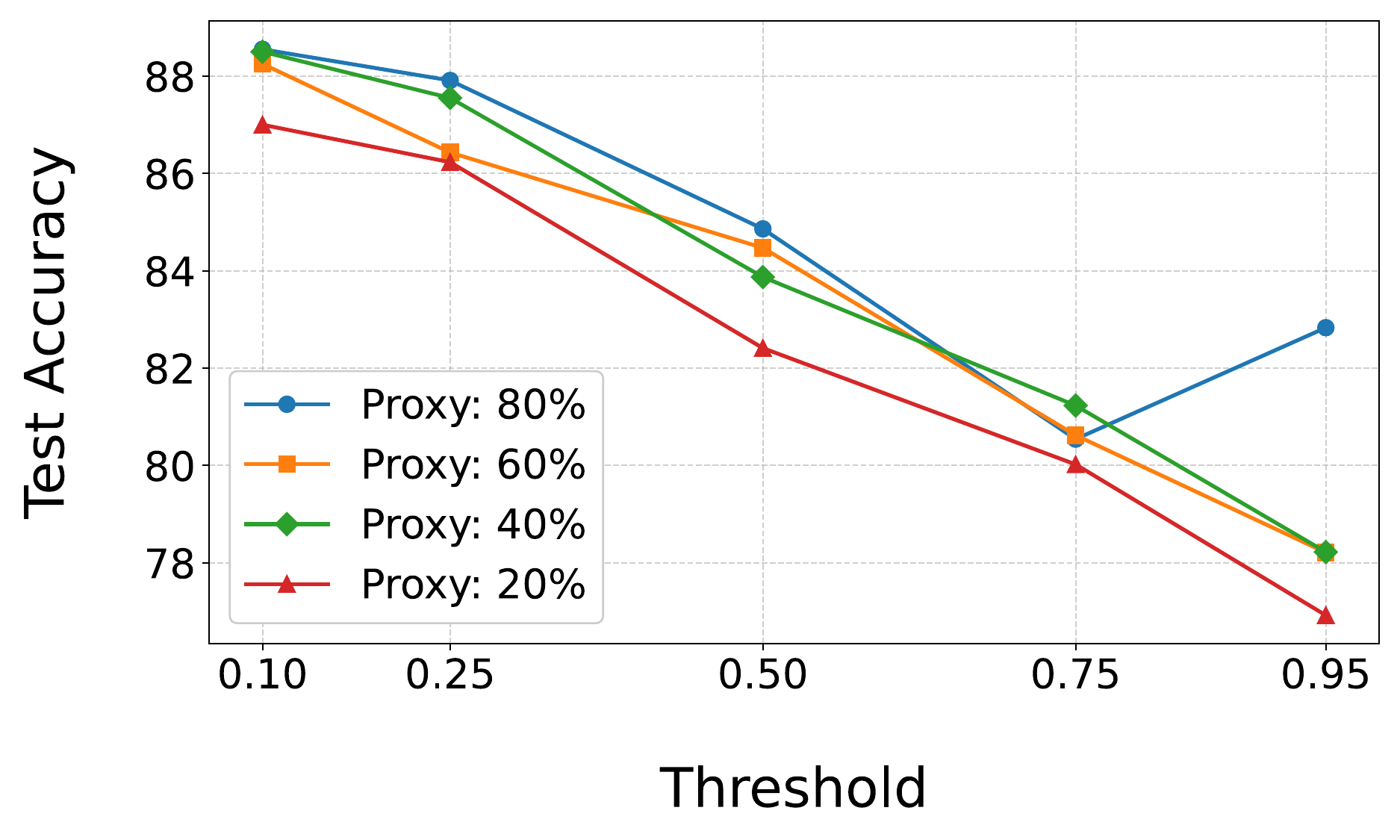}
        \caption{FashionMNIST.}
        \label{fig:fashionmnist_proxy}
    \end{subfigure}
    \hfill
    \begin{subfigure}[b]{0.32\textwidth}
        \centering
        \includegraphics[width=\columnwidth]{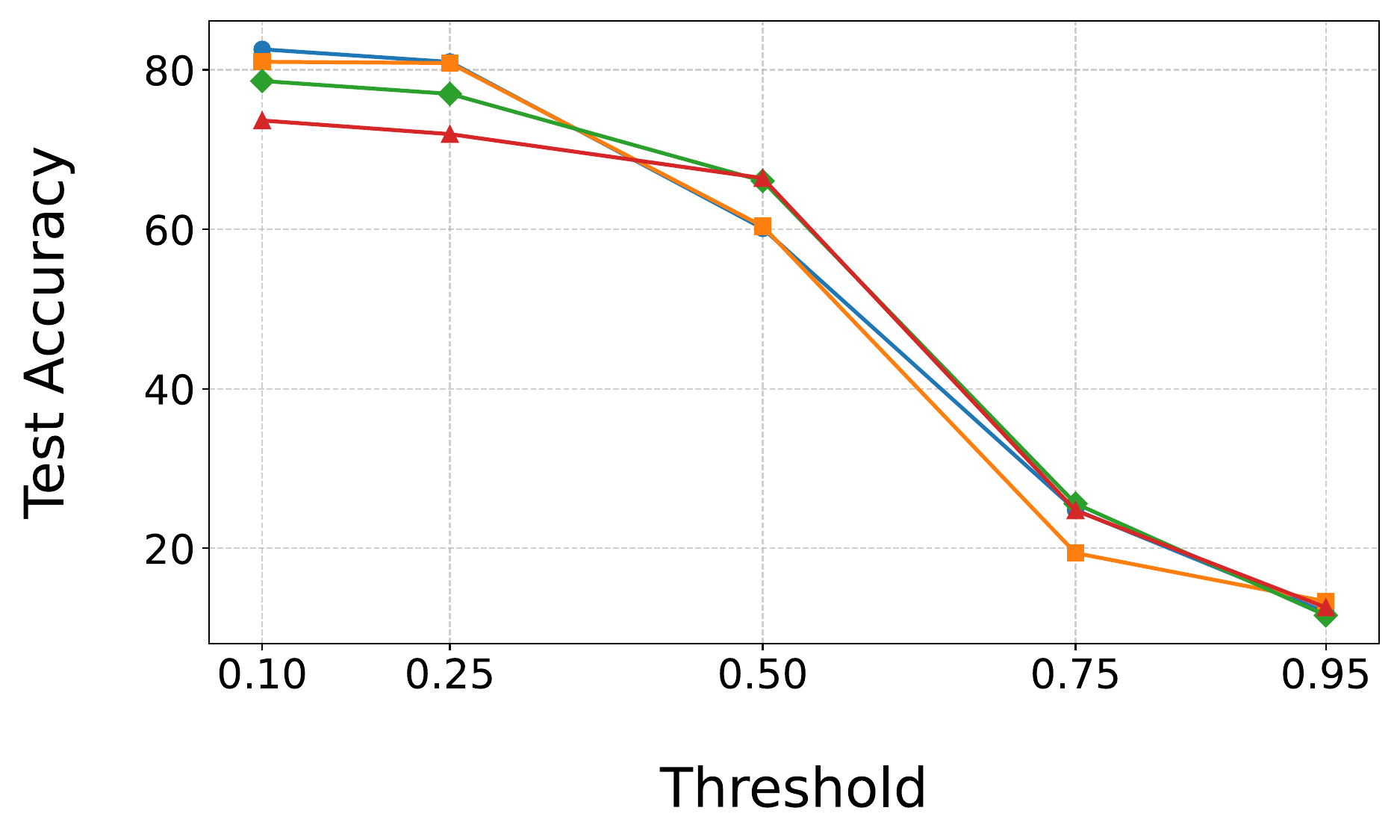}
        \caption{CIFAR10*.}
        \label{fig:cifar10_proxy}
    \end{subfigure}
    \caption{Effect of proxy samples percentage and ID detection threshold on test accuracy for different datasets.}
    \label{fig:merged_figure}
\end{figure*}

\subsubsection{Dataset distribution}
MNIST and FashionMNIST contain grayscale images with distinct class clusters (see Fig.~\ref{fig:mnist} and Fig.~\ref{fig:fashionmnist}), which support effective learning and estimation of DREs. However, complex datasets like CIFAR10 containing three-channel information and overlapping high-dimensional class distributions (see Fig.~\ref{fig:cifar10}), limit its performance. SoTA methods~\cite{shao2024selective} use pre-trained models like ImageNet to extract low-dimensional features for learning DREs on complex datasets. In our implementation, we extract low-dimensional features from CIFAR10 using ImageNet-pretrained on ResNet-18, removing its last layer. These low-dimensional meaningful features enable DREs to determine densities more effectively and improve ID and OOD sample filtration (see Fig~\ref{fig:cifar10_clusters}).

\subsubsection{Threshold}
Client-side threshold regulates the filtering of ID and OOD proxy samples by KMeans-DRE and significantly impacts the performance of EdgeFD. As presented in Fig.~\ref{fig:mnist_proxy} and Fig.~\ref{fig:fashionmnist_proxy}, higher thresholds correlate with decreased test accuracy, increasing OOD sample inclusion in ID proxy data, and corrupting the global convergence in FD. This effect intensifies for CIFAR10 due to inter-class feature overlap (see Fig.~\ref{fig:cifar10_proxy}). Even with low-dimensional feature extraction, larger thresholds lead KMeans-DRE to misclassify OOD samples as ID samples, degrading the final performance of EdgeFD.

\subsubsection{Proxy data}
Using proxy data for the exchange of knowledge among clients is a common practice in feature-based FD methods\cite{jeong2018communication, shao2024selective, FedMD}. Clients share fractions of their private data as a proxy for knowledge distillation. Analysis across all datasets (Fig.~\ref{fig:mnist_proxy}, Fig.~\ref{fig:fashionmnist_proxy}, and Fig~\ref{fig:cifar10_proxy}) reveals that increasing private data requirement from 20\% to 80\% yields minimal performance gains in EdgeFD, thanks to our robust KMeans-DRE and the two-staged client-side filtering strategy. Concretely, EdgeFD achieves effective knowledge distillation with only 20\% of the client's private data.

\subsection{Privacy Leakage in Proxy Data}
Two privacy threats exist in feature-based FD methods.
\paragraph{Pre-distillation leakage} arises before the FD process when generating and sharing the proxy dataset among the clients and the server, posing serious privacy threats by sharing a small percentage of the client’s private data;
\paragraph{Post-distillation leakage} occurs during the distillation rounds, when exchanging the model logits as ``dark knowledge'', leading to significantly lower privacy leakage than sharing the model parameters in FL \cite{shao2024selective}.

Several countermeasures can reduce pre-distillation leakage without eliminating the benefits of proxy data, which remain as part of future work:
\begin{itemize}
\item Share non-sensitive samples or generate synthetic ones, preserving the data distribution without revealing actual private information;
\item Apply noise to the proxy data before sharing, adjusted based on the sensitivity of the underlying data, providing mathematical guarantees against reconstruction attacks at the cost of reduced model accuracy;
\item Employ secure multiparty computation or homomorphic encryption with increased overhead, enabling predictions on proxy data without directly exposing private data.
\end{itemize}

\subsection{Limitations and Future Work}
The current  EdgeFD implementation faces three key limitations for edge computing applications, subject to future work.
\paragraph{Non-uniform client  data} Real-world scenarios involve varying data-sharing due to privacy concerns, data availability, and bandwidth constraints. This imbalance in proxy data can bias distilled local models towards data-rich clients.
\paragraph{Complex datasets} While EdgeFD demonstrates superior performance on the simpler datasets, complex datasets require more efficient feature extractors due to overlapping high-dimensional class distributions.
\paragraph{Proxy data sharing} Alternative approaches like synthetic data generation or differential privacy mechanisms could potentially eliminate the need for direct proxy data sharing, though potentially at the cost of performance. 

\section{Conclusion}
\label{sec:conclusion}
We presented EdgeFD, a novel federated distillation method that addresses critical challenges of data heterogeneity and resource constraints in edge-based federated environments. EdgeFD introduces two significant innovations: a resource-efficient KMeans-DRE approach that captures data distribution through centroid positions rather than complex matrix operations, and a two-stage client-side filtering strategy that eliminates the need for server-side selection while effectively distinguishing between in-distribution and out-of-distribution proxy data. The KMeans-DRE implementation shows linear time complexity instead of the exponential growth seen in statistical DRE methods, enabling efficient deployment on devices with limited processing capabilities. Furthermore, EdgeFD achieves effective knowledge distillation with minimal proxy data requirements (20\% of private data), while maintaining performance comparable to scenarios using much larger proxy datasets. EdgeFD's ability to maintain high performance while significantly reducing computational overhead makes it particularly suitable for resource-constrained edge devices. Experimental results demonstrate its superior accuracy over SoTA FD methods across heterogeneous data distributions while achieving effective generalization with minimal proxy data. This work advances scalable collaborative learning on edge devices, bridging the gap between centralized AI and decentralized applications.

\section*{Acknowledgments}
This work received funding from the European Union MSCA COFUND project CRYSTALLINE (grant agreement 101126571), the ``University SAL Labs'' initiative of Silicon Austria Labs (SAL) and its Austrian partner universities for applied fundamental research for electronic-based systems, and the  Austrian Research Promotion Agency (FFG grant agreement 909989 ``AIM AT Stiftungsprofessur für Edge AI'').

\bibliographystyle{IEEEtran}
\bibliography{references}

\end{document}